\documentclass{article}

\usepackage{PRIMEarxiv}

\usepackage[utf8]{inputenc} 
\usepackage[T1]{fontenc}    
\usepackage{hyperref}       
\usepackage{url}            
\usepackage{booktabs}       
\usepackage{amsfonts}       
\usepackage{nicefrac}       
\usepackage{microtype}      
\usepackage{lipsum}
\usepackage{amsmath}
\usepackage{graphicx}
\graphicspath{{media/}}     
\usepackage{xcolor}
  
\title{MoWE : A Mixture of Weather Experts}

\author{
  Dibyajyoti Chakraborty, Romit Maulik\\
  The Pennsylvania State University \\
  State College, PA 16801, United States\\
  \texttt{d.chakraborty@psu.edu} \\
   \And
  Peter Harrington, Dallas Foster, Mohammad Amin Nabian, Sanjay Choudhry \\
  NVIDIA Corporation\\
  Santa Clara, CA 95051, United States  \\
}

\begin{document}
\maketitle

\begin{abstract}
Data-driven weather models have recently achieved state-of-the-art performance, yet progress has plateaued in recent years. This paper introduces a Mixture of Experts (MoWE) approach as a novel paradigm to overcome these limitations, not by creating a new forecaster, but by optimally combining the outputs of existing models. The MoWE model is trained with significantly lower computational resources than the individual experts. Our model employs a Vision Transformer-based gating network that dynamically learns to weight the contributions of multiple "expert" models at each grid point, conditioned on forecast lead time. This approach creates a synthesized deterministic forecast that is more accurate than any individual component in terms of Root Mean Squared Error (RMSE). Our results demonstrate the effectiveness of this method, achieving up to a 10\% lower RMSE than the best-performing AI weather model on a 2-day forecast horizon, significantly outperforming individual experts as well as a simple average across experts. This work presents a computationally efficient and scalable strategy to push the state of the art in data-driven weather prediction by making the most out of leading high-quality forecast models.
\end{abstract}


\section{Introduction}
The discipline of global medium-range weather forecasting has been recently redefined by a wave of data-driven models, which have achieved unprecedented performance in terms of forecast skill and speed. Foundational architectures like FourCastNet \cite{pathak2022fourcastnet}, Pangu-Weather \cite{bi2022pangu}, and GraphCast \cite{lam2023learning} established milestones by demonstrating superior accuracy in deterministic metrics compared to leading physics-based Numerical Weather Prediction (NWP) systems. This initial success, driven by deep learning on massive reanalysis datasets, represented a remarkable leap forward, creating forecasts at a fraction of the computational cost of traditional methods. In the years since, a large number of data-driven (and hybrid physics-ML) models have been developed, building on early successes and further advancing the state-of-the-art \cite{bodnar2025foundation,nguyen2024scaling,chen2023fengwu,chen2023fuxi,kochkov2024neural,price2023gencast,lang2024aifs,bonev2025fourcastnet,willard2025analyzing}.

With the proliferation of data-driven weather forecast models, the variations in model architecture, training objective and recipe, number of variables, resolution of the training data, and finetuning curriculum has similarly seen a rapid increase, and thus a diverse set of behaviors is observed in forecasts produced by these models. To address this, the research community has taken on the challenge of analysis and intercomparison between models, investigating their forecast skill as a function of variable and lead time \cite{rasp2024weatherbench}, their physical properties \cite{bonavita2024some}, their capacity to represent extremes \cite{bonavita2024some,ben2024rise}, their long-term stability \cite{karlbauer2024advancing,watt2023ace,mccabe2023towards}, and their behavior in ensemble settings \cite{brenowitz2025practical,mahesh2024huge,price2023gencast,alet2025skillful}, among other more event-specific evaluations like tropical cyclones \cite{demaria2025operations} or heat waves \cite{vonich2024predictability}.

While distinct behaviors, advantages, and disadvantages of specific models have now been well-identified, the overall trajectory of progress in terms of forecast skill has begun to slow and many models attain similar performance on bulk metrics, especially in a deterministic forecast setting \cite{rasp2024weatherbench}. This convergence in performance suggests an emerging possibility: rather than any single architecture consistently outperforming others across all variables, regions, and lead times, different models may excel in different circumstances, and combining the strengths of multiple models could yield forecasts superior to any single contributor.

In the numerical weather prediction (NWP) community, multi-model and ensemble approaches have long leveraged diversity among forecasts to improve skill and reliability \cite{buizza2018development,park2008tigge}. Multi-model approaches have also been explored in the realm of data-driven weather models but in a more limited sense -- most works have focused on multi-model ensembling within a given architecture class \cite{weyn2019can,kochkov2024neural,alet2025skillful,mahesh2024huge}, often training the same model from different random seed initializations as a means of producing more model variability in ensemble forecasts. While promising, such an approach may fail to fully exploit the heterogeneous capabilities of modern AI weather models. Only recently have works explored more diverse ensembles incorporating multiple source models \cite{liu2024evaluation,doe2025piggycast}, but these have only been explored at very coarse resolution or with simpler ensembles that just average across several source models.

In this work, we present a Mixture of Experts (MoWE) framework that directly addresses this gap and explores the possibilities of combining existing state-of-the-art models. Rather than building a standalone forecaster, our approach learns to optimally fuse the predictions of multiple pre-trained expert models at each grid point, conditioned on lead time. A lightweight transformer-based gating network assigns spatially and temporally varying weights to each expert’s output, enabling the synthesis of a deterministic forecast that selectively draws on strengths of the source models. This dynamic, fine-grained blending strategy produces a coherent global forecast that consistently surpasses the skill of the best single expert and significantly outperforms naive averaging, offering a novel path forward in the increasingly saturated landscape of data-driven weather prediction.

The key contributions of this paper are as follows:
\begin{enumerate}
    \item \textbf{A novel MoWE framework for AI weather forecasting} that dynamically combines the outputs of multiple state-of-the-art models at each grid point and lead time, rather than relying on a single model or static ensemble weights.

    \item \textbf{A lightweight transformer-based gating network} that learns spatiotemporal adaptive expert weights from forecast fields which can be rolled out in an autoregressive manner alongside the expert models to make  superior forecasts.

    \item \textbf{Empirical evaluation and ablation studies} on global medium-range forecasts, demonstrating consistent RMSE improvements of up to 10\% over the best-performing individual expert at a 2-day lead, and substantial gains over simple averaging.
    
\end{enumerate}

\begin{figure}
    \centering
    \includegraphics[width=0.8\linewidth]{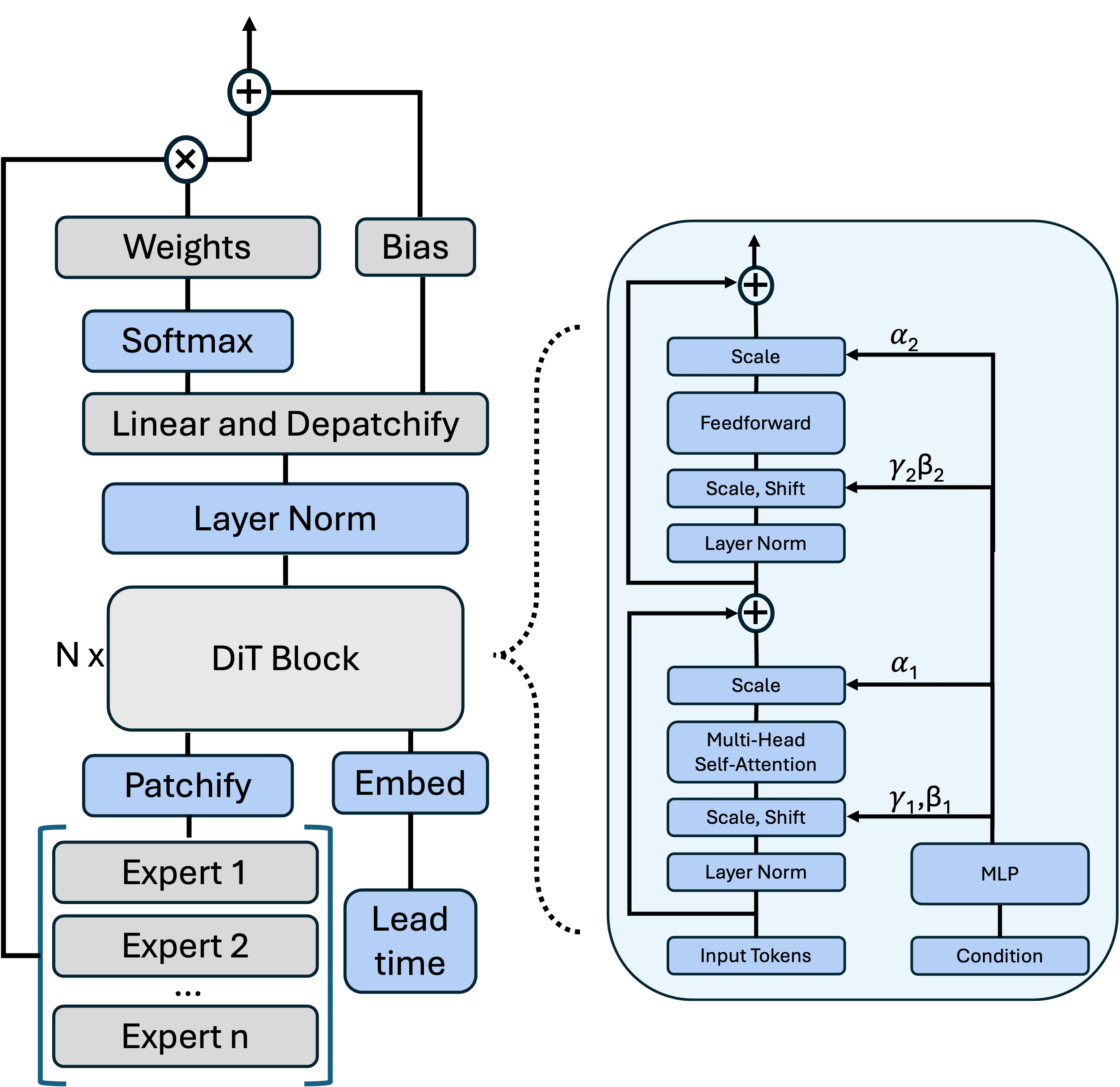}
    \caption{A diagram illustrating the architecture of the Mixture-of-Weather Experts (MoWE). The left panel shows the overall data flow, where inputs from multiple experts are concatenated and patched, with lead time embeddings, and processed through a series of N processing blocks. The right panel provides a detailed view of a single block. These blocks are same as a Diffusion Transformer(DiT) block \cite{peebles2023scalable} with core component: a multi-head self-attention modulated by conditioning information through adaptive scaling and shifting parameters.}
    \label{fig:placeholder}
\end{figure}


\section{Methodology}
The Mixture of Experts (MoWE) model is designed to produce a superior forecast by combining the outputs of multiple, pre-existing "expert" models. Instead of choosing a single best model, it learns to dynamically weight the contributions of each expert based on the specific conditions of the forecast. The core of this system is a gating network, which is a deep neural network architecture that determines these optimal weights.

The fundamental premise is that for a given set of $N$ expert models, each producing a forecast $E_i$, we can construct a more accurate, synthesized forecast $\hat{Y}$ through a spatitemporally varying weighted combination. The governing equation for this synthesis is:
\begin{equation}\label{eq:mowe}
     \hat{Y} = \sum_{i=1}^{N} (W_i \odot E_i) + b 
\end{equation}

The final prediction, $\hat{Y}$, is created by blending the forecasts from several individual ``expert'' models, $E_i$. For each expert, the model learns a ``weight map,'' $W_i$, that assigns a specific level of importance to that expert at every single point on the grid. After multiplying each expert's forecast by its corresponding weight map (the $\odot$ symbol), a final ``bias map,'' $b$, is added to correct any systematic errors shared by all the models.

We employ a gating type network,  $f_{\text{gate}}$, that learns to produce these weights and the bias term by considering all expert forecasts as well as additional conditioning information:

$$ (W_1, W_2, \dots, W_N, b) = f_{\text{gate}}(E_1, E_2, \dots, E_N, t, z) $$

Here, $t$ represents the forecast lead time, and $z$ is an optional noise vector used in the probabilistic variant of the model to capture forecast uncertainty by generating an ensemble of predictions.
The model's architecture heavily utilizes Vision Transformer (ViT) \cite{dosovitskiy2020image} blocks in the "gating network." It works by first stacking the forecast maps from all the different expert models into a single, multi-channel image. This composite image is then broken into patches and fed through the Transformer layers. Crucially, the network's processing at each step is dynamically adjusted based on the forecast's lead time and an optional noise vector using adaptive layer norm. After processing, the model doesn't output a weather map directly. Instead, it outputs a set of pixel-by-pixel "weight maps"—one for each expert—and a final bias map. A softmax layer is employed at the end to have a unit sum of the weights at each point on the grid, per channel. These weights specify how to blend the expert forecasts at every point in the grid to create the final, improved prediction.

In this preliminary study, we use three expert models to provide source forecasts for our MoWE:  Pangu \cite{bi2022pangu}, Aurora \cite{bodnar2025foundation}, and FCN3 \cite{bonev2025fourcastnet}. The Pangu model uses an innovative approach treating the Earth's atmosphere as a three-dimensional data cube, allowing the model to capture complex vertical and horizontal weather patterns simultaneously. Its training recipe involves a hierarchical temporal aggregation strategy, where different models are trained for various forecast lead times. In our case we use the mix of 6 and 24 hour Pangu models. Aurora is built on a Swin Transformer with Perceiver-based encoders and decoders, allowing it to process and learn from a diverse range of atmospheric data. Its training involves a pretraining phase on ERA5 and simulation data to learn general atmospheric dynamics, followed by a fine-tuning phase for specific tasks such as high-resolution weather forecasting. This makes it highly competitive with the state-of-the-art in short and medium range weather forecasting. In contrast to Pangu and Aurora, FCN3 uses a spherical neural operator and a hidden Markov model to generate probabilistic forecasts. By sampling different noise realizations, FCN3 produces ensemble forecasts and is trained to minimize the Continuous Ranked Probability Score (CRPS) \cite{lang2024aifs} metric. Since it is a probabilistic model, the single-member scores of FCN3 appear to lag behind Pangu and Aurora in deterministic metrics, but the FCN3 ensemble surpasses the skill of leading conventional ensemble forecasts and rivals the best diffusion-based methods (see \cite{bonev2025fourcastnet} for detailed evaluation). For our purposes of developing an MoE, these models were chosen for their high degree of accuracy, ease of use, alignment on input and output variables and time steps, and to have a variation across architecture, training objective, and deterministic vs. probabilistic forecast skill in the experts. We use the NVIDIA earth2studio\footnote{\texttt{\href{https://github.com/NVIDIA/earth2studio}{https://github.com/NVIDIA/earth2studio}}} implementations for each of the models.

The MoWE training data consists of independently generated 2-day forecast trajectories by each expert model initialized at each timestep of ERA5 data, spanning 1980 to 2014. From this dataset, we sample random initial conditions and random lead times in batches for training. The MoWE is trained to produce the single best forecast by minimizing the Mean Squared Error (MSE) loss between its prediction $\hat{Y}$ and the ground truth $Y$ at the requested lead time. We use the year 2015 for testing. For autoregressive rollouts, we rollout individual models for 2 days independently and then use the MoWE at each timestep to obtain the forecast. For instance, for a forecast at 12 hour, we take the autoregressive forecasts for each expert at 12 hour and input them in the MoWE model along with the lead time. The MoWE model gives the respective weights for each expert and a bias which is used to generate the MoWE forecast using Equation \ref{eq:mowe}. We open-source our code and release it through NVIDIA PhysicsNeMo\footnote{\texttt{\href{https://github.com/NVIDIA/physicsnemo/tree/main/examples/weather/mixture_of_experts}{https://github.com/NVIDIA/physicsnemo/examples/weather/mixture\_of\_experts}}}.

\section{Results}
\begin{figure}[!ht]
    \centering
    \includegraphics[width=0.99\linewidth]{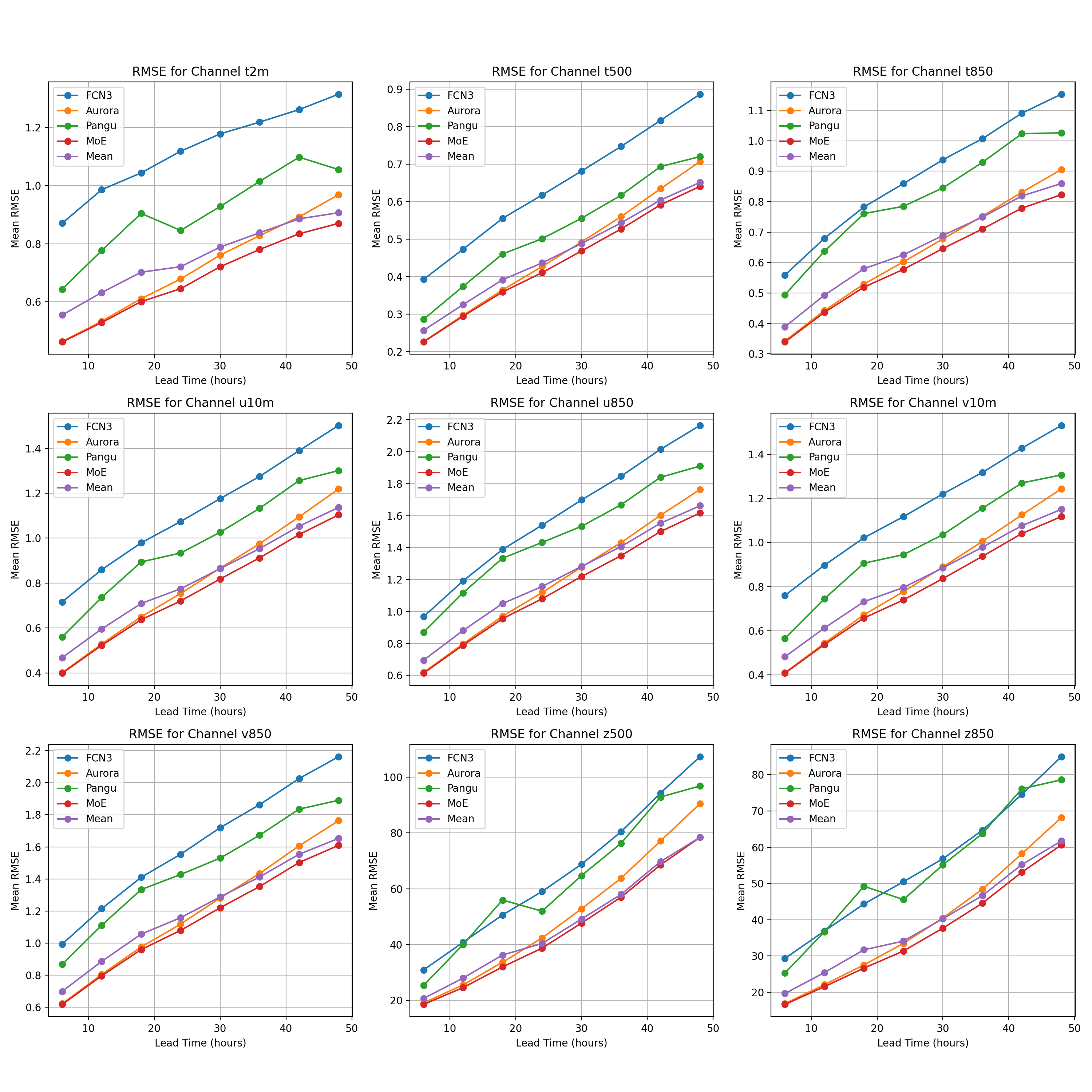}
    \caption{RMSE Comparison for Weather Forecasting Models. The Root Mean Squared Error (RMSE) is plotted against the forecast lead time (in hours) for nine different atmospheric variables. Our trained Mixture of Experts (MoWE) model is compared against three individual expert models (FCN3, Aurora, Pangu) and a simple mean of the experts. Across all variables and lead times, the MoWE model (red line) consistently achieves the lowest error, outperforming not only each expert but also the simple mean model.}
    \label{fig:rmse}
\end{figure}

\begin{figure}[!ht]
    \centering
    \vspace{-0.5cm}
    \includegraphics[width=0.91\linewidth]{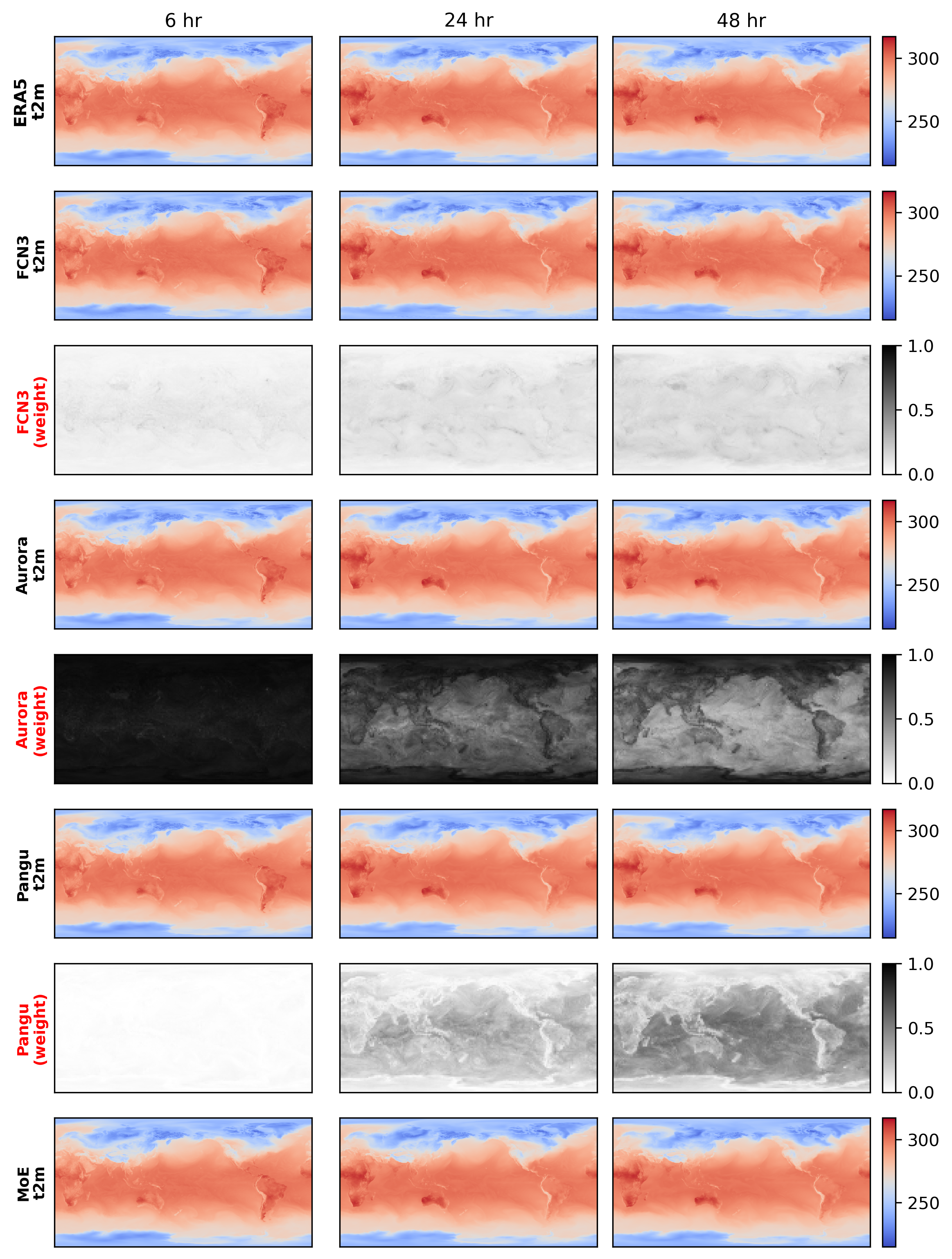}
    \caption{This image displays 2 m temperature (K) for ERA5, forecasts at 6-hour, 24-hour, and 48-hour from various models ( FCN3, Aurora, Pangu, and MoWE) along with the learned weights for FCN3, Aurora, and Pangu in the MoWE. The MoWE model's forecasts appear reasonable and visually consistent with the others. The weight maps indicate that at the 6-hour forecast, the MoWE places a higher weight on the Aurora model. However, as the forecast horizon extends to 24 and 48 hours, the MoWE uses a balanced contribution from multiple models in the longer-term forecasts.}
    \label{fig:t2m}
\end{figure}

\begin{figure}[!ht]
    \centering
    \vspace{-0.5cm}
    \includegraphics[width=0.91\linewidth]{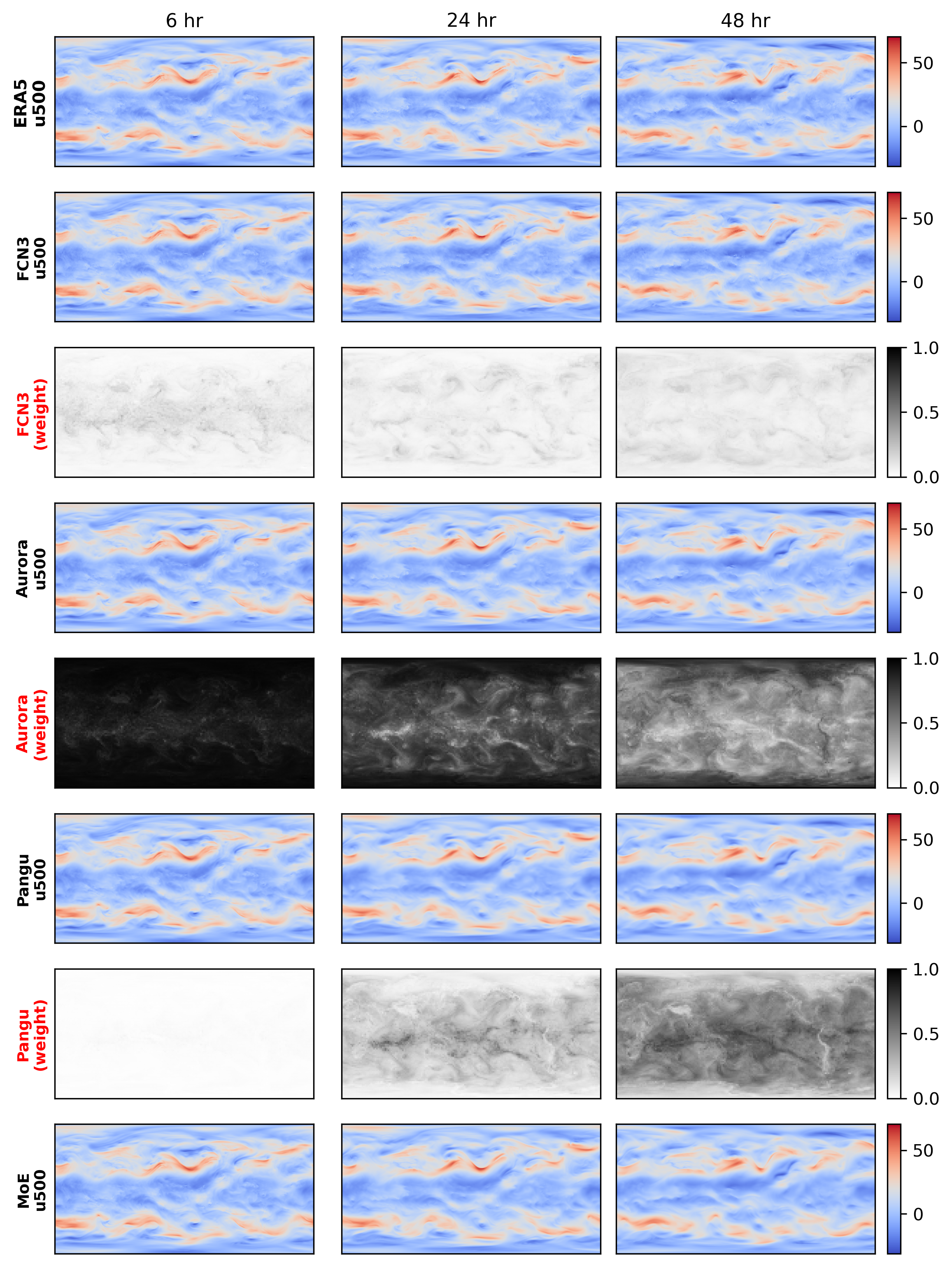}
    \caption{This image displays wind($m/s$) at pressure level 500 for ERA5, forecasts at 6-hour, 24-hour, and 48-hour from various models ( FCN3, Aurora, Pangu, and MoWE) along with the learned weights for FCN3, Aurora, and Pangu in the MoWE. The MoWE model's forecasts appear reasonable and visually consistent with the others. The weight maps indicate that at the 6-hour forecast, the MoWE places a higher weight on the Aurora model. However, as the forecast horizon extends to 24 and 48 hours, the MoWE uses a balanced contribution from multiple models in the longer-term forecasts.}
    \label{fig:u500}
\end{figure}

\begin{figure}[!ht]
    \centering
    \vspace{-0.5cm}
    \includegraphics[width=0.91\linewidth]{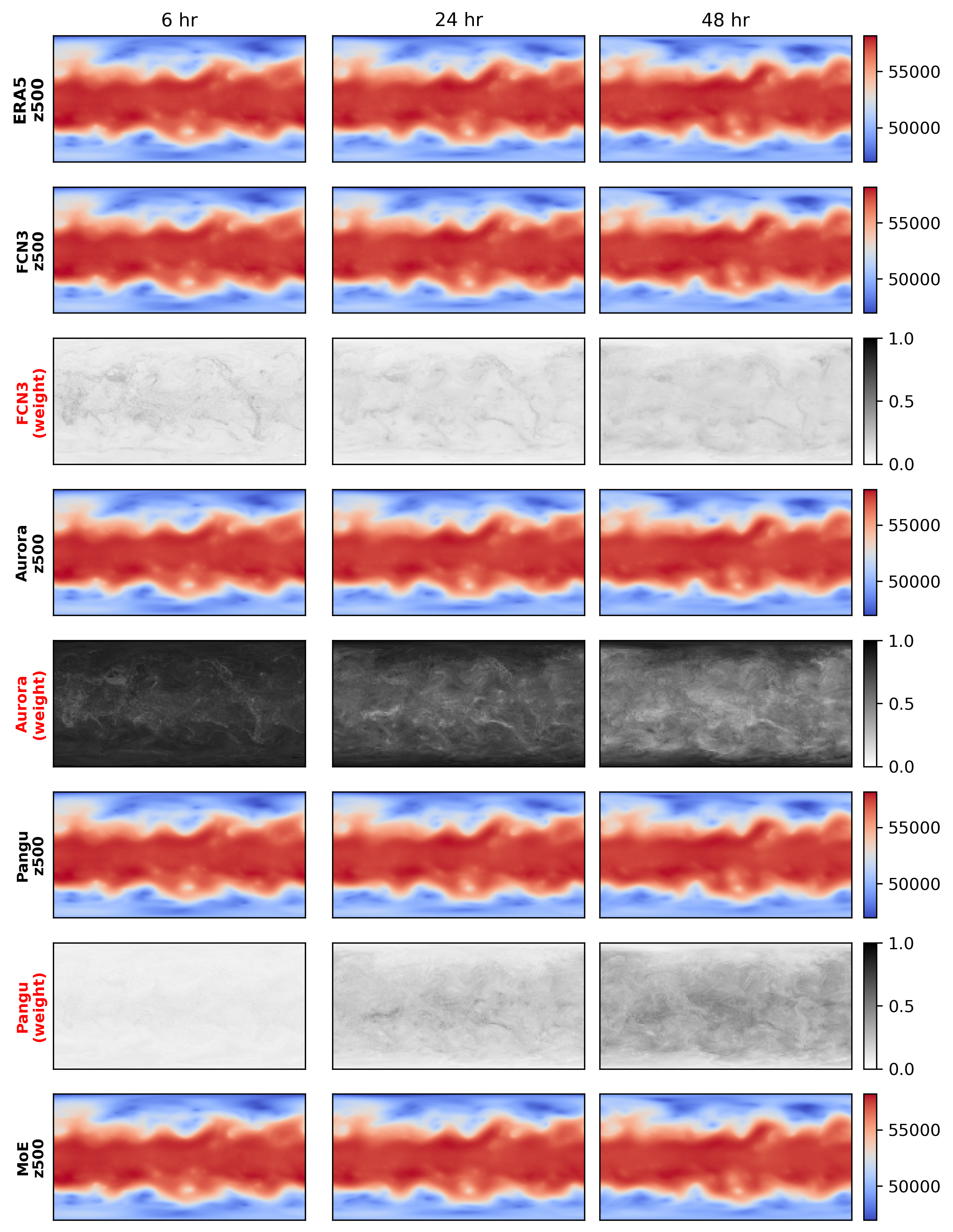}
    \caption{This image displays geopotential ($m^2/s^2$) at pressure level 500 for ERA5, forecasts at 6-hour, 24-hour, and 48-hour from various models ( FCN3, Aurora, Pangu, and MoWE) along with the learned weights for FCN3, Aurora, and Pangu in the MoWE. The MoWE model's forecasts appear reasonable and visually consistent with the others. The weight maps indicate that at the 6-hour forecast, the MoWE places a higher weight on the Aurora model. However, as the forecast horizon extends to 24 and 48 hours, the MoWE uses a balanced contribution from multiple models in the longer-term forecasts.}
    \label{fig:z500}
\end{figure}

We train the Mixture of Weather Experts (MoWE) to combine forecasts from the three expert weather models across various lead times, from 6 hours up to 2 days. We also compare it with a basic mean "mixture" model that simply takes the mean of all experts as the prediction. The mean model is actually a relatively challenging baseline, because it outperforms individual experts at longer lead times. Close to the initial condition of a forecast, individual experts perform better due to higher predictability, but by around 1-2 day lead-time, the mean model becomes superior as averaging reduces error, effectively producing an ensemble-mean-like solution. This transition highlights an interesting dynamic and tradeoff in forecast skill as measure by RMSE \cite{brenowitz2025practical}. Any successful mixture of experts approach should be able to improve on both the best individual expert as well as the simple mean across experts, and this is indeed what is observed in Figure \ref{fig:rmse}. For the most challenging 2-day forecast horizon, the MoWE model achieves a Root Mean Squared Error (RMSE) that is 10\% better than the score of the best individual expert model, and we observe the MoWE outperforming the mean mixture at all variables and lead times evaluated.

To investigate the impact of MoWE model capacity, we perform a minimal ablation study across model sizes. We train a Base model (25 million parameters) and a Small model (9 million parameters) with the same architecture, using the hyperparameters in Table \ref{Hyperparam_table}, the results of which are listed in Table \ref{ablation}.

\begin{table}[!ht]
\centering
\caption{Hyperparameters for Base (25M) and Small (9M) models.}
\begin{tabular}{lcc}
\hline
\textbf{Hyperparameter} & \textbf{Base} & \textbf{Small} \\
\hline
Patch size                & 8           & 8 \\
Hidden size               & 384         & 256 \\
Depth (layers)            & 6           & 3 \\
Attention heads           & 6           & 4 \\
MLP ratio                 & 4.0         & 4.0 \\
\hline
\end{tabular}
\label{Hyperparam_table}
\end{table}

\begin{table}[!ht]
\centering
\caption{RMSE comparison for Base (25M) and Small (9M) models across different channels, along with percentage difference. The Base model is very slightly improving over the Small model.}
\begin{tabular}{lccc}
\hline
\textbf{Variable} & \textbf{Base} & \textbf{Small} & \textbf{\% Difference} \\
\hline
t2m (K)   & 0.6803  & 0.6810  & +0.10 \\
t500 (K) & 0.4399  & 0.4410  & +0.27 \\
t850 (K) & 0.6038  & 0.6046  & +0.12 \\
u10m ($m/s$) & 0.7663  & 0.7664  & +0.01 \\
u500 ($m/s$) & 1.4872  & 1.4875  & +0.02 \\
u850 ($m/s$) & 1.1404  & 1.1405  & +0.01 \\
v10m ($m/s$) & 0.7844  & 0.7844  & +0.01 \\
v500 ($m/s$) & 1.5019  & 1.5023  & +0.03 \\
v850 ($m/s$) & 1.1420  & 1.1417  & -0.03 \\
z500 ($m^2/s^2$) & 45.6734 & 45.9691 & +0.65 \\
z850 ($m^2/s^2$) & 36.5431 & 36.8664 & +0.88 \\
\hline
\end{tabular}
\label{ablation}
\end{table}

We observe from Table \ref{ablation} that the Base model performs better than the Small model, though the difference is marginal. Attaining such good performance even with a lightweight model showcases the effectiveness of our MoWE framework in utilizing the pretrained expert models efficiently. We use the Base model hereon for the results.

The MoWE model generates forecasts that are qualitatively consistent with baseline models, as shown in the Figures \ref{fig:t2m},\ref{fig:u500}, and \ref{fig:z500}. In addition, the weights show the ability to be dynamically adjusted based on lead time, channels, and spatial locations. Initially, at the 6-hour forecast, the MoWE heavily favors the Aurora model as it is the most accurate. Although FCN3 appears to perform poorly in our comparison, it is important to note that we are evaluating a single ensemble member of FCN3. Individual members of a stochastic model generally exhibits higher RMSE, which is consistent with its intended nature. As the forecast progresses to 24 and 48 hours, the weights are distributed more evenly among the constituent models (FCN3, Aurora, and Pangu). In particular, the spatial patterns of these weights are not arbitrary; they appear to be influenced by geographical features such as coastlines and continents. This suggests that the MoWE is learning a physically relevant, spatiotemporally dependent strategy for combining the different models' outputs.

\section{Conclusion}
In this work, we addressed the emerging performance saturation in deterministic data-driven weather forecasting by proposing a Mixture of Experts (MoWE) framework as a strategic alternative to training yet another standalone model. Our approach leverages the strengths of multiple existing expert models, using a learned gating network to produce an optimal, dynamic blend of their forecasts.

Our results show that this synergistic approach is highly effective in improving forecast skill. By combining outputs from three distinct deep learning models, our MoWE system produced forecasts that were quantitatively superior to any single expert, achieving a significant RMSE reduction of over 10\% at a 48-hour lead time. This demonstrates that there is complementary and valuable information distributed across different models that can be effectively harnessed. Moreover, the superiority of the MoWE against simpler ensembling strategies (e.g., averaging) shows that the advantages of different experts between models can be isolated to specific locations and lead times. Thus our approach could also be considered a novel way to approach the problem of bias correction in weather forecasting.

We acknowledge some of the weaknesses of our MoWE architecture and training procedure. First, our current approach fixes the rollout to 2 days, as we precompute and store all expert forecasts on disk to cut down on training time. A potential solution to this (and the relatively large size of the training datasets in general) is an online training setup, where the expert predictions are computed on-the-fly during training. This would demand substantially more GPU memory and/or longer training times, so we leave it to future work. Another limitation is that simple concatenation of all channels across experts becomes increasingly infeasible as the number of variables and experts grows. This highlights the need for compression or dimensionality reduction strategies, for example, using a perceiver IO \cite{jaegle2021perceiver} to compress the data across experts, channels, or pressure levels \cite{bodnar2025foundation}.

Overall, the implications of this research offer a promising path forward for the discipline. As the number and diversity of leading weather models continues to grow, the MoWE framework provides a scalable and efficient method to maximize forecast accuracy by integrating the collective intelligence of the best available models, without adding substantial extra compute cost on the training or inference sides. Future work could extend this framework to include a more diverse set of experts, including traditional physics-based NWP systems, and investigate the learned weights to gain insight into model-specific strengths and weaknesses, among other potential directions. Ultimately, we see the MoWE approach as a shift from competing models to collaborative models, paving the way to the next generation of weather forecasting systems with community effort and involvement.

\section*{Acknowledgements}
D. Chakraborty acknowledges PhysicsNeMo and Earth2Studio teams at NVIDIA. This research was done during his internship at NVIDIA. R. Maulik and D. Chakraborty acknowledge funding support
from the DOE Office of Science ASCR program (DOE-FOA-2493, PM-Dr. Steve Lee).

\bibliographystyle{unsrt}  
\bibliography{references}

\end{document}